\documentclass{article}

\usepackage[preprint]{neurips_2026}

\usepackage[utf8]{inputenc} 
\usepackage[T1]{fontenc}    
\usepackage{hyperref}       
\usepackage{url}            
\usepackage{booktabs}       
\usepackage{amsfonts}       
\usepackage{nicefrac}       
\usepackage{microtype}      
\usepackage{xcolor}         
\usepackage{amsmath} 

\usepackage{amsthm}

\usepackage{amssymb}   
\usepackage{amsfonts}

\theoremstyle{definition}
\newtheorem{definition}{Definition}

\usepackage{epigraph}

\usepackage{tcolorbox}
\tcbuselibrary{skins}

\definecolor{MyDeepBlue}{RGB}{0, 51, 102}
\definecolor{MyBoxGray}{RGB}{235, 235, 235}
\definecolor{MyLightBlue}{RGB}{235, 245, 255}

\newtcolorbox{finding}[1]{%
    enhanced,
    frame hidden,
    colback=MyLightBlue,
    colframe=black!70,
    sharp corners,
    boxsep=0pt,            
    top=2mm,               %
    bottom=1mm,            %
    left=2mm,              %
    right=2mm,             %
    middle=0.5mm,
    fonttitle=\small\bfseries,
    attach boxed title to top left={xshift=2mm, yshift=-1.5mm}, 
    boxed title style={
        colback=MyDeepBlue, 
        sharp corners, 
        boxrule=0.pt,
        top=0.mm,         %
        bottom=0.mm       %
    },
    title={\textbf{#1}},
    before skip=3pt,       
    after skip=3pt         
}

\usepackage{subcaption}
\usepackage{graphicx}
\usepackage{booktabs, xcolor, colortbl}

\title{NICE FACT: Diagnosing and Calibrating VLMs in Quantitative Reasoning for Kinematic Physics}

\author{
  \textbf{Jian Lan$^{1,2}$, Zhicheng Liu$^{1}$, Xinpeng Wang$^{1,2,\dagger}$, Yuhao Zhou$^{3}$,
  Haokun Chen$^{1,2,\dagger}$,} \\ \textbf{Jiancheng Lv$^{3}$, Barbara Plank$^{1,2}$, Thomas Seidl$^{1,2}$} \\
  $^1$University of Munich (LMU), Germany $^2$Munich Center of Machine Learning$^3$ Sichuan University\\
  $\dagger$: \text{work done in LMU, now at Meta}
}

\begin{document}
\maketitle

\vspace{-0.5em}
\begin{center}
  \centering
  \includegraphics[width=\linewidth]{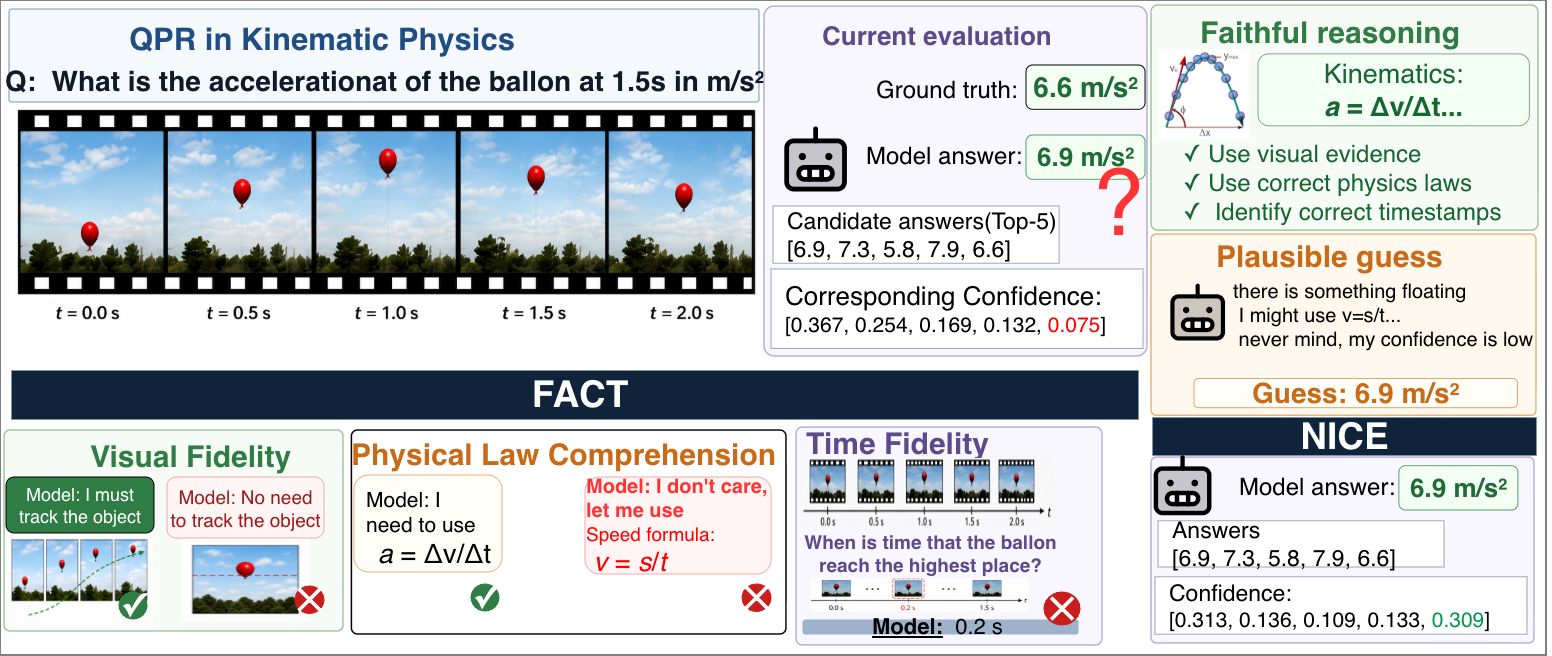}
  \captionof{figure}{Illustration of Qualitative Reasoning in Kinematic Physics. From top-left to the middle, it shows an input and the current evaluation, where MRA fails to reflect either failure reasons or what VLM truly knows: 6.9 seems close to ground-truth but  model pays barely attention to 6.6. This leaves an open question: Is the answer plausible or faithful? MRA alone cannot tell WHY. We use FACT to diagnose why correct/wrong, and use NICE to measure how reliable VLMs are.} 
  \label{fig:main}
\end{center}
\begin{abstract}
The ability to derive precise spatial and physical insights is a cornerstone of vision-language models (VLMs), yet their poor performances in related spatial intelligence tasks such as physical reasoning remain a fundamental barrier. The community critically lacks a scientific analysis revealing whether VLMs \textbf{faithfully} reach answers or \textbf{plausibly} make guesses. This work aims to provide a fundamental understanding of \textit{how VLMs perceive the physical world, and utilize physical laws, while assessing the reliability of model confidence.} We propose \texttt{NICE} and \texttt{FACT}, a dual-diagnostic paradigm that explicitly decomposes quantitative reasoning for kinematic physics. \texttt{FACT} 
diagnoses visual \textbf{f}idelity, physic\textbf{a}l law \textbf{c}omprehension, and \textbf{t}emporal grounding. \texttt{NICE} studies our novel neighborhood-informed calibration method and novel metrics to evaluate and calibrate confidence reliability. Evaluated across 6 latest state-of-the-art VLMs, we uncover that models fail to identify visual preconditions or utilize necessary physical laws to reach answers. This work highlight and establish a standardized diagnostic paradigm to guide the development of faithful, physically-grounded VLMs.
\end{abstract}
\section{Introduction}
\label{intro}
\vspace{-0.5em}
\begin{center} 
  \textit{\small ``What we observe is not nature itself, but nature exposed to our method of questioning.'' }\\[0.2em]
  \hfill \textit{\small --- Werner Heisenberg (1958), Nobel Laureate in Physics}
\end{center}
\vspace{-0.5em}
Physical Reasoning (PR) is a cornerstone of next-generation Vision-Language Models (VLMs)~\citep{qwen3vl,llava-next}, as it enables models to evolve toward deeper understanding of the real physical world’s underlying rules, with the ultimate goal of interacting with humans using physics knowledge~\citep{quantiphy25,chow2025physbench}. Previous works have focused on qualitative assessments, evaluating whether VLMs possess physical commonsense such as identifying static object size using multiple choice question (MCQ)~\citep{star}. As the field matures, the demand for VLMs has transitioned from such mere qualitative intuition to more challenging quantitative physical reasoning (QPR) task such as Kinematic Physics, which requires models generating numeric answers (GNA)~\citep{wu2025indoor,cambrian}. QPR exposes critical vulnerabilities in VLMs as they still lag far behind humans with poor performances. However, current evaluation paradigms face a fundamental limitation, where they predominantly rely on accuracy metric Mean Relative Accuracy (MRA)~\citep{quantiphy25}, which provide a holistic score but offer no insight into the cognitive parts where model fails, or confidence distribution indicating model uncertainty. Although we can observe a model's struggle, the underlying `why' remains hidden. We use Fig.~\ref{fig:main} for illustration.

Without a fundamental understanding of how VLMs perceive the physical world and utilize physical laws, efficiently enhancing VLMs remains elusive. There is an urgent need for a diagnose paradigm that explicitly disentangle these intertwined factors and provide a granular roadmap for building more faithful VLMs. To this end, we first raise 2 core research questions: \textbf{RQ1:} Do VLMs identify, comprehend, and locate the minimal cognitive dimensions: visual, physical law-based, and temporal preconditions necessary for reasoning as humans do? \textbf{RQ2:} Are VLMs reliably confident in their answers with the awareness of ground-truth boundaries, and how to reach a better calibration?

We address these questions by proposing the first dual-diagnostic paradigm for QPR in Kinematic Physics. Our core insight is anchored in human cognitive archetypes \citep{lake2017building}, where we novelly propose that: \textit{a human-like model must manifest a synchronized reasoning flow, where it first reads the query, and then identifies minimal necessary preconditions, and finally reason to reach answer while yielding a reliable confidence}. Therefore, we introduce two decoupled yet complementary axes: \texttt{FACT} and \texttt{NICE}. FACT decomposes the reasoning process into 4 interpretable dimensions that mirror human intuitive physics and spatiotemporal awareness: visual \textbf{f}idelity, physical l\textbf{a}w \textbf{c}omprehension, and \textbf{t}emporal grounding. \texttt{NICE} is a calibration method  studying the importance of \textbf{n}eighborhood-\textbf{i}nformed \textbf{c}onfidenc\textbf{e} distribution. We also propose 2 novel calibration metrics to better support evaluation. Our paradigm moves beyond monolithic accuracy metrics toward a mechanistic understanding of VLM behavior. We reveal 3 counter-intuitive findings: (i) VLMs' failures are rooted in fundamental perceptual inaccuracies (ii) VLMs fails to internalize physical laws and is incapable of executing formula-based reasoning for kinematic physics. (iii) The model lacks neighborhood awareness in confidence distribution, failing to identify valid answer boundaries. Critically, our results demonstrate that diagnosing these two axes jointly, yet independently, is essential for building physically grounded and trustworthy vision-language models.

\textbf{Our contributions are:} (1) We propose \texttt{FACT} and \texttt{NICE}, the first systematic diagnostic framework that successfully disentangle how and why VLMs fail in QPR in kinematic physics. FACT shifts the evaluation paradigm from monolithic MRA to mechanistic failure attribution, allowing for the isolation of specific cognitive bottlenecks.
(2) We are the firs ct to identify VLMs' mis-calibrated confidence distribution in Kinematic Physics. We propose 2 novel calibration metrics and 1 novel calibration techniques for quantifying the neighborhood awareness.
(3) With a surge of benchmarks emerging, we reveal a timely shift toward incorporating intermediate diagnostic signals to explicitly evaluate disentangled physical perception cognition. This establishes a concrete, actionable roadmap for diagnosis-driven architectural design, data curation, and training objectives for future development.

\section{Related Work}
\label{sec2}
\textbf{VQA-based PR and Emerging Benchmarks.}
VLM-based PR primarily adopt Visual Question Answering \citep{VQA} as a task format, leveraging QA to manifest underlying reasoning logic (note the scope is totally different from 3D video understanding~\citep{chen2025reasoning}, robotics and world model~\citep{chen2026generativeenginesactionablesimulators}). There is a growing recognition of the importance of PR in VLMs, but the community is only at a start point for this new frontier.~\citet{chow2025physbench} study physical object properties and physics-based dynamics in MCQ manner. Their study remains on qualitative level where the answers are pre-defined and only described in natural language phrase. \citet{vsibench} starts to learn QPR, with a mix of qualitative MCQ. However, they only focus on object size and length estimation and object counting, which are static in image. Recently, \citet{quantiphy25} and \citet{wu2025indoor} propose that QPR in dynamics is a pivotal task scenario. Among the two, \citet{quantiphy25} is the only one providing necessary priors for reasoning, which supports the task more appropriately, while \citet{wu2025indoor} leave the models to purely estimate answers due to the difficulty of gathering prior information. Currently, the evaluation format stays twofold: One paradigm use F1 score for MCQ, while others using MRA~\citep{vsibench} for GNA.

\textbf{Recent Advance and Challenges in QPR.} 
Previously, work like~\citep{spatialvlm,bonnen2024evaluating} emphasize geometric relationships in 3D-scene, which represent only a part of the physical while leave the task stay static. Recently, \citet{quantiphy25} show models fail to elastically adapt reasoning outcomes in response to the scaling of reference objects. More radically, \citet{wu2025indoor} entirely remove the video input, forcing the model to answer solely based on the text, concluding that VLMs' predictions are pre-determined by linguistic biases rather than visual evidence. However, such design merely demonstrates that generative paths inevitably shift when the input space is corrupted while failing to provide a systematic and direct revelation of the underlying linguistic dominance mechanisms. To date, the granularity of diagnosis remains critically under-specified and lacks formal standardization, with most intermediary reasoning processes confined to implicit, black-box interpretations. In contrast, we decompose QPR into surgical-level diagnosis of model failure and reasons. Critically, recent large model studies have highlighted the significance of the model’s confidence distribution~\citep{lan2026,lan2025,lanbias}. To the best of our knowledge, we are the first to identify and address model confidence distribution in QPR.

\section{Problem Statement and Theoretical Formalization}
\label{sec3}

\textbf{The Evaluation Trap.} Given a dataset $\mathcal{D} = \{(\mathcal{V}_i, \mathcal{Q}_i, \mathcal{A}_i)\}_{i=1}^N$, where $\mathcal{V}_i$ is a video, $\mathcal{Q}_i$ a question, and $\mathcal{A}_i$ the ground-truth, a model $\mathcal{M}$ is optimized to find:
$\hat{a}_i = \arg\max_{a \in \mathcal{Y}} P(a \mid \mathcal{V}_i, \mathcal{Q}_i ; \theta)$
where $\mathcal{Y}$ is the answer space. An outcome-oriented metric $\mathcal{S}$ then quantifies the discrepancy between $\hat{a}_i$ and $\mathcal{A}_i$: $\mathcal{S} = \frac{1}{N} \sum_{i=1}^N \mathcal{L}(\hat{a}_i, \mathcal{A}_i)$.
$\mathcal{L}(\cdot, \cdot)$ is instantiated in 2 types: for MCQ, it represents indicator function $\mathbb{1}(\hat{a}_i \neq \mathcal{A}_i)$ for \textit{exact match}; for GNA, it represents a normalized distance $| \hat{a}_i - \mathcal{A}_i | / \mathcal{A}_i$ to account for absolute deviations in physical variables (e.g., velocity). The internal trajectory of how a model traverses from raw ($\mathcal{V}_i, \mathcal{Q}_i$) to $\hat{a}_i$ remains \textit{inaccessible}. Consequently, VLM vulnerabilities, particularly their specific physical knowledge, remain under-explored. More critically, currently no work has evaluated VLMs' uncertainty towards its answer space. An answer becomes indistinguishable between two mechanisms, where previous benchmarks fail to disentangle:
\textbf{Faithful Reasoning ($R_f$):} VLMs successfully ground the minimal necessary preconditions, and executes the correct laws to derive the answer, with reliable confidence. 
\textbf{Plausible Guess ($G_p$):} VLMs bypass the minimal necessary preconditions by exploiting text shortcuts, dataset biases, or physical commonsense in pre-trained weights.
Mathematically, the observed success probability is:
\begin{equation}
P(\hat{a} = \mathcal{A}) = P(R_f) + P(\neg R_f)P(G_p \mid \neg R_f).    
\end{equation}
\textbf{Formalizing the Diagnostic Decomposition and Theoretical Rationale.} 
Following the anchored human cognitive archetypes in Sec.~\ref{intro}, we define (also see~\ref{app:theo} for more details):
\begin{definition}[Cognitive Necessity]
Let $\mathcal{R}$ be a reasoning trace generated by a model, and $\Phi = \{\phi_1, \dots, \phi_n\}$ be the set of minimal necessary cognitive conditions (e.g., visual information, physical laws). We define $\mathcal{R}$ as \textbf{faithful} if and only if:
\begin{equation}
    \forall \phi_i \in \Phi, \quad \mathbb{I}(\mathcal{R} \models \phi_i) = 1
\end{equation}
where $\mathcal{R} \models \phi_i$ denotes that the reasoning trace logically satisfies the physical constraints of condition $\phi_i$. If $\exists \phi_i \in \Phi$ such that $\mathcal{R} \not\models \phi_i$, the reasoning is defined as a \textbf{spurious derivation}, regardless of the correctness of the final numerical answer. To move \textbf{beyond empirical observation} and establish a principled foundation that is \textbf{not} merely a collection of heuristics, we re-formalize VLM reasoning through the lens of \textit{Causal Information Flow} by partitioning the inference process into controllable probes, allowing us to quantify the "net gain" contributed by each component as follows: 
\end{definition}
\begin{definition}
The reasoning quality $\mathcal{R}_{q}$ is defined as an abstract consistency score of a model under a set of multi-modal factor-specific diagnostic probes $\boldsymbol{\delta}$. $\mathcal{R}_{q}$ quantifies the grounding strength by measuring the distributional divergence $\mathbb{D}$ between the original state $\mathcal{I}$ and the probed state ($\mathcal{I}; \boldsymbol{\delta}$):
\begin{equation}
{\mathcal{R}q}_{\Phi} \triangleq \sum_{\phi \in \Phi} \mathbb{E}_{\boldsymbol{\delta}_\phi} \left[ \mathbb{D} \left( \mathcal{M}(\mathcal{I}) \parallel \mathcal{M}(\text{do}(\mathcal{I}; \boldsymbol{\delta}_\phi)) \right) \right]
\label{eq-rq}
\end{equation}
where do($\cdot$) is the do-calculus operation~\citep{Pearl}, which performs a controlled intervention on specific diagnostic factors $\boldsymbol{\delta}_\phi$. The divergence $\mathbb{D}$ thus measures the \textit{causal influence} of each condition $\phi$: a faithful model should capture minimal necessary preconditions when asked by a query. In Sec~\ref{sec4}, we translate $\mathcal{R}_q$, $\delta$, $\mathbb{D}$ and $\Phi$ into measurable targets, which allows us to decompose the monolithic reasoning process into interpretable, quantifiable components of cognitive necessity. The core intuition is to treat reasoning faithfulness as a \textit{causal sensitivity} problem, which quantifies the extent to which the model's reasoning is truly \textbf{grounded} in the necessary cognitive conditions $\Phi$. 
\end{definition}

\section{FACT \& NICE: A Dual Diagnostic Paradigm}
\label{sec4}
\textbf{Bridging Theoretical Foundation and Operational Diagnostics.} While Sec.~\ref{sec3} formalizes solid theoretical foundation, the quantification of $\Phi$ requires a concrete operational vehicle. To bridge this gap, we instantiate a \textit{dual-diagnostic paradigm} that operationalizes each factor through measurable signals. Our approach consists of two complementary component \textbf{FACT} and \textbf{NICE} as follows.

\textbf{FACT.} Following Sec.~\ref{sec3}, we define 3 minimal necessary dimensions in set $\Phi$ that constrain the conditional probability $P(a | \mathcal{V}, \mathcal{Q})$: Visual \textbf{f}idelity (VF,$\phi_F$): models must correctly perceive query-relevant minimal universal preconditions in visual input (details in next paragraph); Physical L\textbf{a}w \textbf{C}omprehension (PLC,$\phi_A$): models must correctly use appropriate governing physical laws; \textbf{T}emporal Grounding (TG,$\phi_T$): models must correctly align reasoning with the temporal structure of the scene. Since we cannot directly observe how VLMs reason across $\Phi$, we then construct \textit{diagnostic probes} that transform the problem into \emph{subtasks}. Formally, a probe $\delta_\phi$ maps an input $(V, Q)$ to a derived task: $(V, Q) \;\xrightarrow{\;\delta_\phi\;}\; (V_\phi, Q_\phi)$ where a model’s response is given by: $\hat{a}_\phi = \mathcal{M}(V_\phi, Q_\phi).$ Crucially, correctness on the probed task serves as an observable proxy: if the model fails on $(V_\phi, Q_\phi)$, it indicates that the required functional dependency on $\phi$ is weak or absent, regardless of its performance on the original task. Unlike direct input perturbations or directly manipulating latent variables, we instead study models' intermediate cognitive process. A model is grounded in factor $\phi$ if its responses remain consistent with the constraints imposed by $\delta_\phi$. We introduce details about $\phi_{F,A,T}$ below. Due to page limits, we only keep key concepts in main text. We explain more with examples in appendix.~\ref{app:annotation}, which show clearly our design is theoretically and scientifically sound.

\texttt{Visual Fidelity (VF)}: To operationalize $\phi_F$, we design a set of Universal Kinematic Descriptors that represent the visual preconditions required to identify when answer any ($\mathcal{Q,V}$) pair. 
We distill four canonical dimensions as options and construct MCQ:
(1) \textit{In-depth information}: Must preconditions linked to depth of field, such as camera distance, be specified? (2) \textit{Multi-frame reasoning}: Is multi-frame tracking needed for tracking object motion? (3) \textit{Event Spotting}: (4) \textit{Scale Reference}: Must a reference object or coordinate scale be identified first? These forces the model to exhibit its perceived fidelity, e.g., if a model cannot identify that "Scale Reference" is required for Newton's Second Law, any subsequent answer is likely to be spurious. We manually annotate the ground truth for each $\mathcal{(Q,V)}$ pair, requiring the model to select all valid patterns. Our 4 descriptors serve as attributes in $\phi_F$, where F1 score serves as measure function $\mathbb{D}$ to get the distance between the 4-dimension vector $v_{\phi_F}$ with the customized MCQ ground-truth. A higher F1 score indicates a model satisfies more necessary minimal preconditions when answering a query.

\texttt{Physical Law Comprehension (PLC)}:
Similar to VF, we distill 5 classical formulas (see~\ref{app:annotation}) as universal minimal necessary precondition attributes in $\phi_A$, and construct MCQ to let a model identify which formulas will be used for a certain query. This decouples law selection from numerical computation, enabling diagnosis of whether errors arise from incorrect reasoning rules rather than arithmetic mistakes. The measurement stay the same as above.

\texttt{Temporal Grounding (TG)}:
To operationalize $\phi_T$, since there is no dataset supporting temporal grounding for kinematic physics reasoning, we annotate a timestamp-event subset in both GNA and MCQ manner for Quantiphy (details in~\ref{app:annotation}). For GNA, we require VLMs to identify the exact timestamp of a given event, evaluated by MRA. For MCQ, we require VLMs to select the correct event from candidate set given a timestamp, evaluated by F1 score. They both test VLMs' temporal indexing precision. Correct responses indicate that the model respects temporal dependencies.

\textbf{NICE}:\textbf{n}eighborhood-\textbf{i}nformed \textbf{c}onfidenc\textbf{e} calibration and evaluation.
A fine-grained understanding of the model's \textbf{confidence level} is important for reliability~\cite{lan2026,lan2025}. This introduces a fourth, orthogonal dimension $\phi_C$ which governs the \textit{confidence distribution} of model outputs. We emphasize it is even more essential for GNA tasks where the ground-truth, despite being a single scalar, resides within a continuous semantic neighborhood where values in certain neighborhood carry nearly identical physical implications. Consequently, this neighborhood must not be neglected. There are 2 typical phenomena: \textbf{(1)} Ideally, a lower confidence in the neighboring values of the ground-truth is justifiable, but only if the model exhibits exceptionally high MRA-Acc, reflecting a state of absolute certainty. \textbf{(2)} However, for a model characterized by epistemic uncertainty, the probability mass should ideally manifest as a unimodal distribution (e.g., Gaussian-like) concentrated around the top-tier candidates. A sharp, "cliff-like" drop-off in confidence—where only a single value receives high probability while its immediate semantic neighbors are assigned near-zero mass—is symptomatic of poor calibration and a lack of robustness in the model's underlying reasoning space. Since GNA tasks operate in an open-ended generative space, they lack a gold distribution typically found in closed-set classification, which makes the traditional divergence-based metrics such as Kullback-Leibler (KL) Divergence \citep{kl} or Expected Calibration Error (ECE) \citep{guo2017calibration} inherently inapplicable. To fill this gap, we introduce \texttt{Nicon}, a novel \textbf{N}eighborhood-\textbf{i}nformed \textbf{c}alibrati\textbf{on} method and 2 novel calibration metrics. Given a neighborhood candidate set $\mathcal{{N}_c}$= $[c_1,\cdots,c_T]$, the ground truth $\mathcal{A}_i$, and number of neighbors $T$, we propose: 

\textbf{GT Neighborhood Confusion Index (NCI).} $NCI$ measures the probability distribution near $\mathcal{A}_i$. We uniformly sample $T$ points $\{\hat{c}_t\}_{t=1}^T$ within the range $[\mathcal{A}_i - \Delta, \mathcal{A}_i + \Delta]$ with a fixed step size:
\begin{equation}
    NCI(\mathcal{A}_i; \Delta, T) = \frac{\sum_{t=1}^{T} P(\hat{c}_t \mid \hat{c}_t \in [\mathcal{A}_i - \Delta, \mathcal{A}_i + \Delta], \hat{c}_t \neq \mathcal{A}_i)}{T \cdot P(\mathcal{A}_i)}, \hat{c}_t = \mathcal{A}_i - \Delta + (t-1) \frac{2\Delta}{T-1}
\label{eq4}
\end{equation}
A high $NCI$ indicates that the model is certain that answer should be very close to the ground-truth within the range $[\mathcal{A}_i - \Delta, \mathcal{A}_i + \Delta]$, whereas $NCI \approx 1$ reveals a desired distribution where VLMs have similar confidence with in the data points in the range, and a low NCI means the model does not have neighborhood awareness near ground-truth, regardless of accuracy.

\textbf{Neighborhood Calibration Error (NCE).} 
Unlike traditional calibration metrics \citep{stop}, NCE explicitly accounts for the distance between $\hat{c}_t$ and $\mathcal{A}_i$, with $\hat{c}_t$'s confidence as weight:
\begin{equation}
NCE =  \frac{\sum_{t=1}^{T} P(c_t) \cdot \Psi(c_t, \mathcal{A}_i)}{T}, \Psi(c_t, \mathcal{A}_i) = \max \left( \zeta, \text{ } 1 - \frac{|c_t - \mathcal{A}_i|} {|\mathcal{A}_i|+\epsilon} \right)
\end{equation}
where $\Psi(c_t, \mathcal{A}_i)$ is the Physical Alignment Weight. $\zeta$ is a lower-bound truncation constant (we set $\zeta = -1, \epsilon=0.001$) to make $\Psi(c_t, \mathcal{A}_i) \in$ [-1,1]. In essence, $NCE$ ensures that the score is maximized only if the model's probability mass is precisely concentrated on the high-fidelity neighborhood of the ground truth. It treats confidence as a "weighted vote": a vote for a precise answer is a reward, while a high-confidence vote for an erroneous answer is a severe penalty. 

\textbf{Nicon.} We further propose \textbf{N}eighborhood-\textbf{I}nformed \textbf{C}alibrati\textbf{on} (Nicon) to help calibration since there is no previous work studying this. Nicon penalizes "isolated spikes" in the probability space and rewards candidates with robust local support. We first define the Local Consistency Factor $\rho(v)$ as the ratio of the average neighborhood probability to the anchor point's probability. We use beam search to generate different possible answers, constructing generation candidate set $\mathcal{N}_{bs}$, $|\mathcal{N}_{bs}|=K$. For each value $u_k$ in $\mathcal{N}_{bs}$, we extract its neighborhood set $\mathcal{N}_{u_k}$, $|\mathcal{N}_{u_k}|=K$. 
\begin{equation}\rho(u_k) = \frac{\sum{P(u), u\in \mathcal{N}_{u_k}}} {K \cdot P(u_k)}, \tilde{P}(u_k) = \frac{1}{Z} \left[ P(u_k)^\alpha \cdot \rho(u_k)^{(1-\alpha)} \right], Z = \sum_{j=1}^{K}\left[ P(u_k)^\alpha \cdot \rho(u_k)^{(1-\alpha)} \right]
\end{equation}
$\rho(u_k)$ is a probability density function that serves as a scaling mechanism. The extent of this scaling is \textbf{contingent upon} the local density of the possible answers. The philosophy is build based on each possible answer's neighborhood awareness. If a value has better neighborhood awareness distribution, that means the model is more certain in this range, which should be amplified, and vice versa. $Z$ serves as normalization function. $\tilde{P}(v)$ is the Calibrated Confidence that represents the refined probability mass after considering neighborhood consistency. $alpha$ balances the model's raw prediction against the neighborhood constraint, where $\alpha \in [0, 1]$ is the trust weight. We emphasize that this mechanism acts as a \textbf{discriminative filter}: for "strong" models with high $NCI$, the anchoring facilitates a reliable consensus among proximal candidates; for "weak" models lacking neighborhood awareness, the calibrated confidence will naturally collapse, revealing the model's underlying epistemic limitations. 

All together, we define the full factor set: $\Phi_t = \{\phi_F, \phi_A, \phi_C, \phi_T\}$
where $[\phi_F, \phi_A, \phi_T]$ captures reasoning overall fidelity and $\phi_C$ captures confidence reliability. This decomposition is minimal in the sense that each component corresponds to a necessary and functionally distinct aspect of QPR, and cannot be trivially removed without losing explanatory power. Due to page limits, we use~\ref{app:theo} to help explain more foundations that supports our design.

\section{Experiments}
\subsection{Setup Details.} 
We choose 6 latest SOTA VLMs: Qwen3VL-8B \citep{qwen3vl}, Molmo2-8B \citep{molmo2}, InternVL3.5-8B \citep{internvl35}, Gemma3-12B \citep{gemma3}, Qwen3-Omni-30B \citep{qwen3omni}, and GLM4.6V-9B \citep{glm}, with a design where the first three are vision-language dual-encoder models, while the latter three are unified models. The selection is standardized and same with the work~\citep{quantiphy25}. Note that we only use open-sourced models as studying internal mechanism such as confidence distribution and time fidelity. It is impossible to study these for close-sourced models. We have covered reasonably sufficient VLMs compared with previous works~\citep{wu2025indoor}. We also exclude much weaker models like Phi-4 \citep{phi4}, SmolVLM \citep{smolvlm}, and LLaVAs \cite{llava-next, llava-video}, but provide their results and more reasons in~\ref{more}. To provide fair evaluation, we follow ~\citep{quantiphy25} using the complete open-sourced set from Quantiphy~\citep{quantiphy25}. To the best of our knowledge, Quantiphy is the only open-source dataset that supports quantitative reasoning with preconditions on QPR. Other datasets' limitation are listed in Sec.~\ref{sec2}, which render them inapplicable to our work. All implementation details are in~\ref{app:implementation details}. The setup is crystal clear and highly reproducible. For metrics, we project $\mathbb{D}$ in Eq.~\ref{eq-rq} into a multi-faceted metric space $\mathcal{S} = \{F_1, \mathcal{C}, MRA\}$, where each metric serves a distinct diagnostic purpose: Macro $F_1$ score measures the external performance and robustness of MCQs under perturbation. Confidence $\mathcal{C}$ reflects the model's confidence distribution across candidate answer space. MRA evaluates the model's precision in numeric values. Equations are in~\ref{app:implementation details}.
\subsection{Results and Analysis}

\begin{figure}[t]
  \centering
  \includegraphics[width=\linewidth]{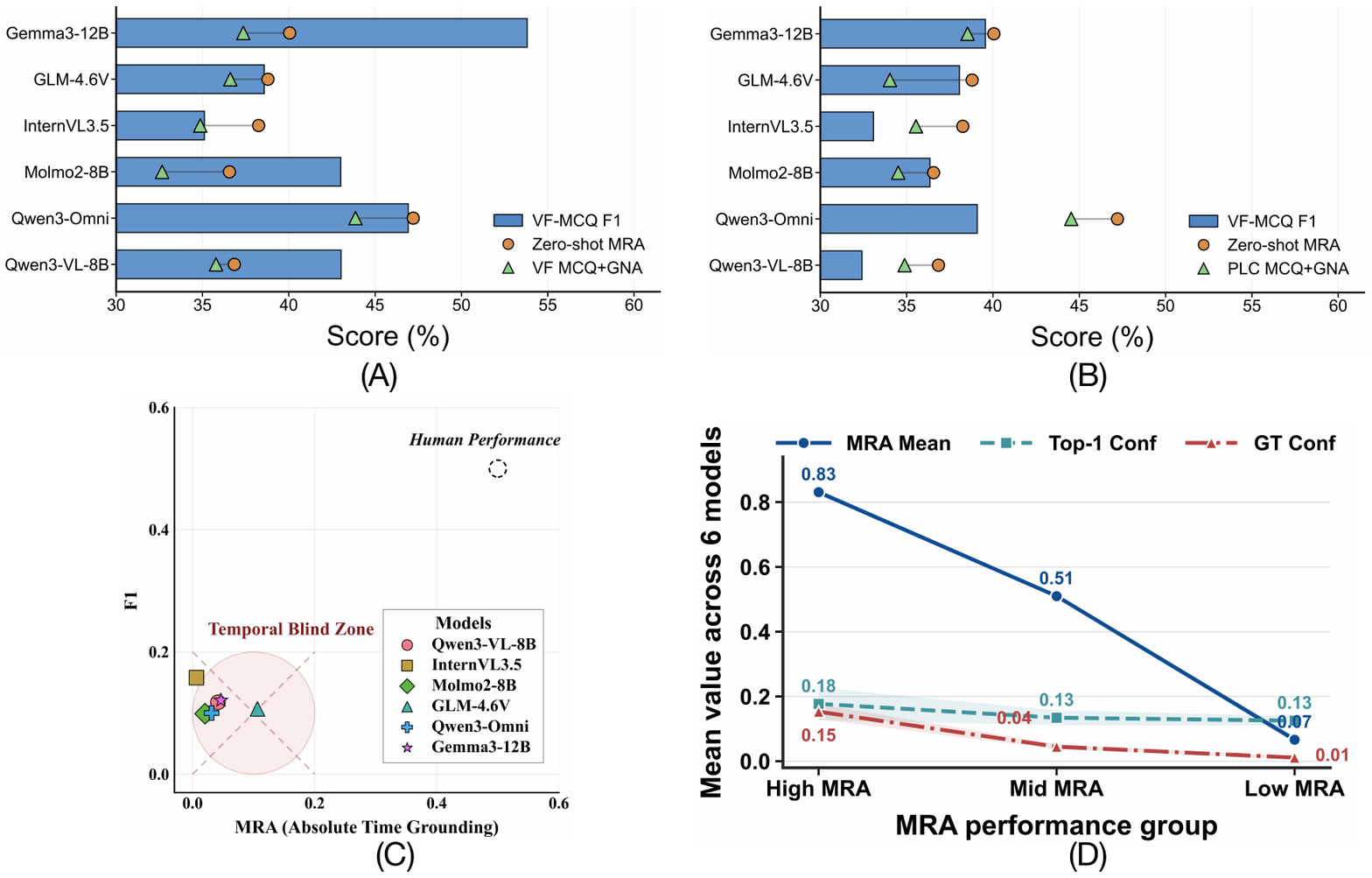}
  \vspace{-0.5em}
  \caption{(A): Performances on visual fidelity (VF). (B): Performances on physical law comprehension (PLC). (C): Performances on Temporal Grounding (TG). (D): MRA vs Confidence.}
  \label{fig2}
\end{figure}

\begin{finding}{Key Finding 1: Visual Perceptual "Blindness" to Critical preconditions.}As shown in Fig.~\ref{fig2}A, VLMs struggle to identify the necessary visual preconditions, yielding a low macro F1-score ($<50\%$, only Gemma3 reach 55\%) on MCQ when identifying Universal Kinematic Descriptors (blue bars). Meanwhile, after wrongly identifying visual preconditions, we prompt VLMs to continue finishing GNA, using MCQ and its own answer as extra Chain-of-Thought (triangles). MRA degrades compared with the zero-shot performances (circles).  

\end{finding}
\textbf{Analysis and Takeaways}: Finding 1 indicates that VLMs lack the consciousness to identify visual preconditions, even under explicit questioning and candidate options. This finding demonstrates that VLMs tend to make plausible guess because models fail to meet minimal necessary visual preconditions. Notably, incorporating MCQ answers into a CoT context yielding lower MRA compared to the zero-shot. We believe the reason is when forcing models to following a Cognitive CoT based on already incorrect priors, models lead themselves to reasoning degradation through hallucinated steps. We formalize this as a Cumulative Hallucination effect $P(\hat{a_1} \mid \text{CoT}(\phi_{F}), \mathcal{V}, \mathcal{Q}) > P(\hat{a_0} \mid \text{Vanilla}, \mathcal{V}, \mathcal{Q})$ where CoT acts as a "logical poison" rather than a scaffold. We discuss more about cumulative effects in Sec.~\ref{discussion}. Note that higher F1 does not translate into higher MRA.

\begin{finding}{Key Finding 2: Stochastic Parrots.}As shown in Fig.~\ref{fig2}B, VLMs struggle to identify correct physical laws, with all F1-scores lower than 40\%. When prompt the VLMs to follow their MCQ selection in CoT manner, the MRA drops same as in Finding 1. The F1 scores are poorer compared with in VF. 
\end{finding}
\textbf{Analysis and Takeaways}: Finding 2 indicates that VLMs do not comprehend physical laws. This is crucial as faithful models must reason via such laws. The behavior is more like `stochastic parrots' where models make plausible guesses following the pre-trained inference path while do not know what itself is doing (we discuss more in Sec.~\ref{discussion}). Essentially, compared with VP, the F1 score is much lower, meaning that comprehending and using formulas is much more challenging. The cumulative effects also stand for PLC, which is consistent with VF.

\begin{finding}{Key Finding 3: VLMs are Lost in Time.} As shown in Fig.~\ref{fig2}C, whether predicting timestamps for given events or answering MCQs about events at specific time, the model exhibits low MRA and F1 scores. More importantly, all selected models landed in a tiny close `blind' zone, which is far from human performances. 
\end{finding}
\textbf{Analysis and Takeaways}: 
Finding 3 indicates a lack of sufficient absolute temporal awareness in VLMs. Crucially, without precise temporal grounding, any calculation is performed on hallucinated durations. We further demonstrate that $\phi_T$ is decoupled with $\phi_F$ and $\phi_A$. Improvement to identify timestamps does not correlate with model's inherent perception on VF and PLC (in Sec.~\ref{discussion}).

\begin{finding}{Key Finding 4: MRA is NOT sufficient! Confidence is Essential.}

As shown in Fig.~\ref{fig2}D, by partitioning the dataset into three tiers based on MRA (High (0.66-1), Mid(0.33-0.66), and Low(0-0.33)), we observe that the model's Top-1 confidence scores remain very low across all tiers. Furthermore, regardless of the proximity to the ground-truth, the model consistently assigns disproportionately low confidence to the ground-truth neighborhood. 
\end{finding}
\textbf{Analysis and Takeaways}: Finding 4 suggests that current MRA fails to reflect the model's complete reliability, as high-accuracy predictions do not correlate with confidence in the answer space. VLMs failing to notice Ground-Truth neighborhood indicates that VLMs do not know what possible optimal directions are. We discuss more confidence results in Sec.~\ref{discussion}.

\subsection{Discussion: Mechanistic Insights, Bottlenecks, and Beyond}
\label{discussion}
\begin{figure}[t]
  \centering
  \includegraphics[width=\linewidth]{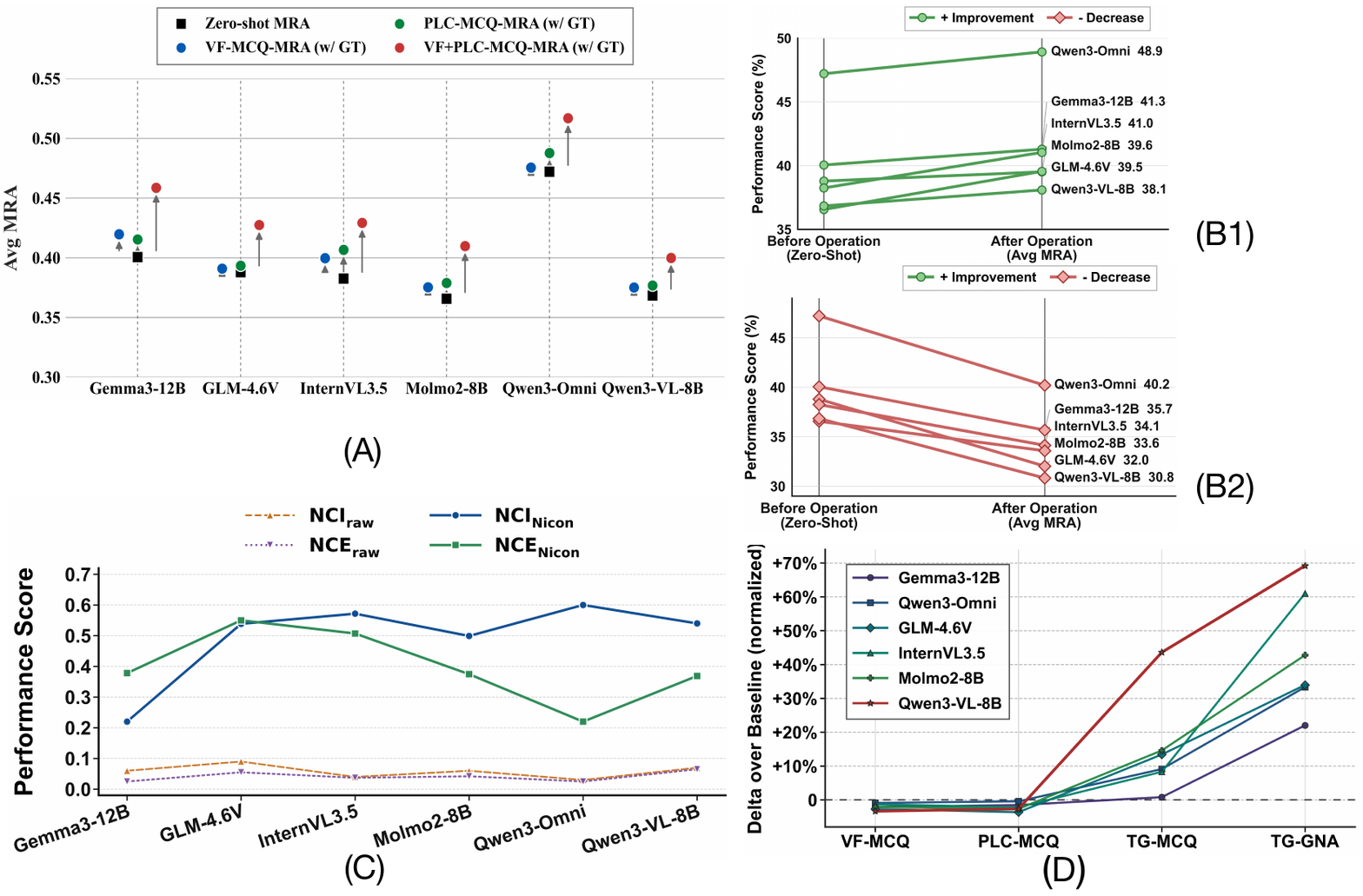}
  \vspace{-0.5em}
  \caption{A: CoT-enhanced MRA with Ground-truth. B1: Replacing PLC options from mathematical laws to formal nomenclature. B2: Add Gaussian noises to visual embeddings. C: NICE results. D: Extra temporal embedding effectiveness.}
  \label{fig3}
\end{figure}

\textbf{Guiding Cognitive Perception.} 
To further verify whether Cognitive CoT truly carries model cognition based on Cumulative Hallucination Effect, we provide the model with corrected MCQ results to re-conduct the reasoning process. As shown in Fig.~\ref{fig3}A, substituting MCQ answers with ground truth improves performance for all models: PLC-MCQ-MRA (use PLC MCQ as CoT to finish MRA) outperforms VF-MCQ-MRA (use VF MCQ as CoT to finish MRA), while the combined VF+PLC MCQ-MRA (use both MCQ) achieves the best results. This not only validates the aforementioned utilizing correct laws is even more essential for our tasks, but also demonstrate the cumulative effect, as it also holds true for the direction when enhanced with ground-truth. We reveal that with correct physical cognitive (a key bottleneck is enforcing models to notice correct preconditions), VLMs are more powerful in QPR in kinematic physics. We highlight that current VLMs indeed do not have awareness to identify physics-related knowledge, and future models should avoid harmful cumulative hallucination effect. 

\textbf{What do VLMs see and read?} A critical open question is: do VLMs truly integrate visual and linguistic cues, or are they biased toward language priors? We argue that previous work do not essentially focus on VLMs' attention on text, neither on query nor formula. Unlike prior works, we propose two criteria for testing whether a model use unknown priors: its behavior should remain consistent with the nature of the applied perturbations. Specifically, for query, replacing a query with a semantically equivalent counterpart should not trigger significant fluctuations in model output. On the vision side, model behavior should fluctuate in direct proportion to the degree of visual interference applied. We apply two semantic perturbations to the query side: 1. substituting mathematical expressions with their technical names, and 2.replacing queries with paraphrased versions. For the vision side, we also employ two methods: 1. adding Gaussian noise to the embeddings; 2. applying perturbations to the raw video at the pixel level. Setting details and more results in~\ref{more}. As shown in Fig.~\ref{fig3}B, it is striking to observe that replacing mathematical formulas with their formal nomenclature leads to an increase in MCQ F1 scores. This indicates that models primarily acquire concept-level knowledge during pre-training, essentially memorizing discrete knowledge concept without a fundamental understanding of their application. On the visual side, the Gaussian noise interference lead to performance degradation. This confirms the model's intent to utilize visual cues, yet the ability is weak: as visual clarity diminishes, the model's vision clue wanes accordingly. We highlight that future work must prioritize advancing models beyond concept-level recognition toward a deeper comprehension of the physical world and the application of physical laws.

\textbf{Confidence Calibration.}
As shown in Fig.~\ref{fig3}C, all VLMs exhibit a severe lack of ground-truth neighborhood awareness and are also mis-calibrated with ground-truth, characterized by near-zero $NCI$ and $NCE$ scores. However, our calibration method Nicon consistently mitigate the issue, improving confidence alignment across all proposed metrics for all models. Additional results regarding traditional calibration metrics such as Temperature Scaling, $\alpha$ ablation study, and discussion on the influence on MRA are provided in the ~\ref{more}. We highlight that future research, particularly within the free-generation paradigm, should prioritize neighborhood awareness.

\textbf{Potential solutions to $\phi_T$.}
We conducted preliminary attempts to enhance the model’s temporal awareness, where we encode temporal information directly into the input tokens. We design a processor-level timestamp injection, which inject per-frame absolute timestamps directly into the token stream at the processor level, without modifying model weights. We adopt two model-specific strategies due to model structure difference. For GLM4.6V and Gemma3, timestamps are injected via interleaved message content: the input sequence is constructed as \texttt{[text `Frame$i$ [t=$t_i$s]:'] [image] [text ``Frame$_{i+1}$ [t=$t_{i+1}$s]:''] [image] $\cdots$}, so that after processor expansion, each timestamp text token block immediately precedes its corresponding visual token block. For the rest, we subclass the respective processor and intervene at the \texttt{<|video\_pad|>} substitution stage: the input\_ids sequence is scanned token-by-token, and for each \texttt{<|vision\_start|>$\cdots$<|vision\_end|>} video block, the corresponding timestamp string (e.g., \texttt{``timestamp: X.XX seconds; ''}) is first encoded into text tokens and then prepended to that block in-place.
In both strategies, timestamps are represented as plain text tokens residing in the same sequence space as visual tokens. As illustrated in Fig.~\ref{fig3}D, this approach yields significant improvements in both MCQ (F1) and GNA (MRA) compared to Finding 4. Building upon this, we further explored whether such performance gains on $\phi_F$ and $\phi_A$. However, Fig.~\ref{fig3}D indicates that such improvement in timestamp identification is insufficient for mitigating VF- and PLC- related issues. This suggests that the absence of temporal awareness is decoupled from the capabilities required for VF and PLC. The performance gap in VF and PLC cannot be bridged by a simple enhancement of temporal consciousness alone. It underscores the urgent need for models with a more robust understanding of the physical world states.

Notably, Gemma 3, Qwen3-Omni, and GLM dominates across different sub-figures on different tasks (except with extra temporal embedding). While no single static model consistently dominates the top rankings, the \textbf{unified model} invariably occupies the superior positions. This trend reveals the potential of unified architectures to serve as a more robust \textbf{development base} for future advancements.

\section{Key Scope, Future and Real-world Impact, and Limitation}
The impact of our work is twofold for the vision-language community: 1. Benchmark Reform: We call for a shift in how to evaluate GNA tasks (QPR) in physical reasoning. Metrics that focus solely on accuracy like MRA may provide a false sense of security regarding a model's general intelligence. We advocate for a more integrated evaluation that couples models' key perceptions and internal mechanism of reasoning, together with confidence distribution. 2. impact on future architecture: our work is the realization that `correcting' the direction of optimization requires more than just post-hoc calibration or temporal token encoding. We have shown that even with improved temporal awareness, models still fail to bridge the gap in physical reasoning. This underscores the need for Robust World Understanding that incorporate inductive physical biases, where true progress in $\phi_F$ and $\phi_A$ will likely require internalizing physical laws (e.g., gravity, causality, and object permanence) during the fundamental pre-training phase, which have a real-world impact for embodied AI~\citep{cambrian}. By identifying this decoupling, we provide a roadmap for the next generation of spatial-temporal VLMs, emphasizing that temporal grounding must serve as a foundation for physical reasoning, rather than an isolated indexing feature. A limitation yet NOT shortcoming is that currently we are not able to expand this work to more datasets because of the lack of data. We are working on constructing new benchmarks following this work and will contribute to the community in the near future. We point out we keep key contents in main text which do not influence the presentation. 

\section{Conclusion}
We introduce FACT and NICE, a dual-diagnostic paradigm that decouples reasoning fidelity from confidence reliability in VLMs for kinematic physics. By transforming minimal necessary cognitive conditions into targeted probes and neighborhood-aware calibration, we shift evaluation from black-box accuracy to mechanistic failure attribution. Our diagnosis reveals cognitive bottlenecks—perceptual blindness, physical law disconnect, weak temporal grounding, and neighborhood-unaware confidence—while NICE calibration effectively mitigates the latter. We contribute and establish a novel standardized, extensible protocol with concrete, actionable roadmap for future architectural design, data curation, and training objectives for faithful and trustworthy VLMs in physical world.
\newpage
\bibliographystyle{unsrtnat} 
\bibliography{references}

\newpage
\appendix

\textbf{Technical appendices and supplementary material}

\section{More Justification for Theoretical Design}
\label{app:theo} 
We justify this diagnostic approach through three theoretical lenses:Causal Identifiability via Intervention: According to the Causal Hierarchy \citep{Pearl}, purely observational data $P(a|\mathcal{V}, \mathcal{Q})$ is insufficient to distinguish between spurious correlations (e.g., language bias) and causal mechanisms (e.g., physical grounding). By applying the $\text{do}(\cdot)$ operator to specific components in $\Phi$, we effectively break the back-door paths from language priors, forcing the model to rely on the remaining information flow.

\paragraph{Minimality and Sufficiency.}
We argue that the triplet $\{\mathrm{VF}, \mathrm{PLC}, \mathrm{TG}\}$ forms a \textit{minimal sufficient set} for reasoning fidelity in quantitative physical reasoning.

\textbf{Sufficiency.}
Given correct visual variable extraction (VF), correct mapping to governing physical laws (PLC), and correct temporal alignment (TG), the model has access to all necessary information required to produce a valid solution for kinematic reasoning tasks. Any correct solution must implicitly satisfy these three conditions, as physical quantities, governing equations, and temporal structure jointly determine the outcome.

\textbf{Minimality.}
Each component is functionally indispensable. 
Without VF, the model lacks correct physical variables, making downstream reasoning ill-posed. 
Without PLC, the model cannot relate variables to the appropriate physical principles, leading to systematically incorrect inference even with perfect perception. 
Without TG, the model fails to align reasoning with the temporal evolution of the system, resulting in violations of causal ordering and dynamic consistency. 
Therefore, removing any component leads to a distinct class of failure that cannot be recovered by the others.

\textbf{Non-redundancy.}
The three factors are mutually non-substitutable: errors in one dimension cannot, in general, be compensated by correctness in others. This ensures that each factor captures a distinct axis of reasoning failure, enabling precise diagnostic attribution.

Together, these properties justify $\{\mathrm{VF}, \mathrm{PLC}, \mathrm{TG}\}$ as a minimal and functionally complete decomposition of reasoning fidelity.

\textbf{Relation to Identifiability.}
This decomposition enables partial identifiability of reasoning mechanisms: by isolating failures along each factor, we can distinguish grounded reasoning from shortcut behaviors that would otherwise be indistinguishable under aggregate performance metrics.

\paragraph{Functional Factorization of Reasoning.}
We further formalize the decomposition $\{\mathrm{VF}, \mathrm{PLC}, \mathrm{TG}\}$ as a structured computation underlying the model’s prediction. Specifically, we view quantitative physical reasoning as a composition of three functional stages:
\begin{equation}
\hat{a} = f_{\mathrm{PLC}} \Big( f_{\mathrm{TG}} \big( f_{\mathrm{VF}}(V, Q) \big) \Big),
\end{equation}
where $f_{\mathrm{VF}}$ extracts task-relevant physical variables from visual input, $f_{\mathrm{TG}}$ aligns these variables with the temporal structure of the scene, and $f_{\mathrm{PLC}}$ maps the resulting representation to the final prediction via governing physical laws.

This formulation does not assume that the model explicitly implements these modules; rather, it provides a \textit{functional abstraction} of the dependencies required for correct reasoning. In particular, each factor corresponds to a necessary transformation whose failure propagates downstream:
errors in $f_{\mathrm{VF}}$ lead to incorrect or missing variables, errors in $f_{\mathrm{TG}}$ lead to misaligned temporal reasoning, and errors in $f_{\mathrm{PLC}}$ lead to incorrect application of physical principles.

\paragraph{Implications for Diagnosis.}
This factorization induces a natural diagnostic strategy: to assess whether a model relies on a given component, we must isolate its functional contribution. However, since these intermediate representations are latent and not directly observable, we cannot access $f_{\mathrm{VF}}, f_{\mathrm{TG}}, f_{\mathrm{PLC}}$ explicitly. 

Instead, we approximate their roles through \textit{factor-targeted diagnostic probes} that selectively test the functional requirements of each stage. If a model’s prediction is insensitive to a probe targeting a specific factor, this suggests that the corresponding functional dependency is weak or absent, indicating a potential shortcut mechanism.

This perspective transforms reasoning evaluation from outcome-based assessment to \textit{structure-aware diagnosis}, where failures can be attributed to specific stages in the underlying computation. In the next section, we operationalize this principle by constructing concrete probe families for each factor, forming the FACT diagnostic framework.

By decomposing the perturbation set $\delta$ into dimension-specific subsets of $\Phi$, we can isolate the information loss within each channel. This allows $\mathcal{R}_q$ to serve not only as a global metric but also as a diagnostic probe to pinpoint which cognitive dimension is the bottleneck of the model's reasoning process. $\mathcal{R}_{q}$ serves as a contour integral over the perturbation on the key dimensions $\Phi$, where any deviation from physical faithfulness manifests as a non-zero distributional drift $\mathbb{D}$ under the do-calculus intervention.

\section{More Justification for Experimental Design, and related details and results}
\label{app:justification-exp}
Same as listed in Sec. 2, world model, robotics, and 3D-video models are excluded as they require extra annotation and processing in bounding-box information, while we focus on end-to-end information flow, which is more natural and similar to the way how humans think. Our work's scope is only with in general zero-shot models, excluding task-specific models with extra post-training. Our aim is to reveal what is important for the next-generation of base-models, rather than task-specific models. Therefore, we do not discuss long-video models which consider complicated methods of processing temporal information.

\subsection{Implementation Details}
\label{app:implementation details}

\begin{table}[h]
\centering
\caption{Core inference and calibration configurations used in our experiments.}
\begin{tabular}{ll}
\toprule
\textbf{Parameter} & \textbf{Value} \\
\midrule
\multicolumn{2}{l}{\textit{Inference Setup}} \\
Hardware & 1 $\times$ H100 80G (single node) \\
Precision & BF16 \\
Decoding & Beam Search \\
\midrule
\multicolumn{2}{l}{\textit{NICE Calibration}} \\
Beam candidates $K$ & 10 \\
Trust weight $\alpha$ & 0.5 \\
\midrule
\multicolumn{2}{l}{\textit{NCI}} \\
Grid interval & 0.5 \\
Neighborhood radius $\Delta$ & 2.5 \\
Grid points $K$ & 10 \\
\midrule
\multicolumn{2}{l}{\textit{NCE}} \\
Grid interval & 0.5 \\
Neighborhood size & 10 (5 each side) \\
Truncation constant $\zeta$ & $-1$ \\
Smoothing term $\epsilon$ & 0.001 \\
\bottomrule
\end{tabular}
\end{table}
We show our hyper-parameters in the above table.
Our experiments are implemented on a single H100 GPU with 80G memory. All the models are directly from open-sourced Huggingface base: https://huggingface.co/.

\textbf{Metrics}
The equation for MRA and KL-divergence is pretty straightforward, and can be easily found in many works~\citep{lan2026,quantiphy25,vsibench}. We show them here again:
KL-Divergence (KL)~\citep{kl}:
\begin{equation}
D_{\text{KL}}( H(s) \parallel P_\mathcal{M}(s) )= \sum_{i=1}^{m} H(s) \log \frac{H(s)}{P_\mathcal{M}(s)}.
\end{equation}
$H(s)$ is human distribution and $P_\mathcal{M}(s)$ is model distribution. the KL score measures how much $H(s)$ deviates from $P_\mathcal{M}(s)$.

To evaluate the model performance more comprehensively, the Mean Relative Accuracy (MRA) is as follows: a discrete set of confidence levels $\mathcal{C} = \{0.1, 0.2, \dots, 0.9, 0.95\}$ is defined. For a given prediction $\hat{y}$ and its corresponding ground truth $y$, the MRA is formulated as:

\begin{equation}
\text{MRA} = \frac{1}{|\mathcal{C}|} \sum_{\theta \in \mathcal{C}} \mathbb{I} \left( \frac{|\hat{y} - y|}{|y|} < 1 - \theta \right),
\end{equation}

\noindent where $\mathbb{I}(\cdot)$ denotes the indicator function, and $|\mathcal{C}| = 10$ represents the number of thresholds. 

\subsection{How we generate model confidence and Temperature Scaling:}
In a Transformer-based architecture, the model generates a sequence of raw score vectors known as \textbf{logits}. To obtain the probability of a specific text sequence, we map these logits to a valid probability distribution. Let $\mathbf{z}_i \in \mathbb{R}^{|V|}$ be the logit vector produced by the model at time step $i$, where $|V|$ is the vocabulary size. The probability of selecting a specific token $x_i$ from the vocabulary is calculated using the \textbf{Softmax} function:
\begin{equation}
P(x_i \mid x_{<i}) = \frac{\exp(z_{i, \text{idx}(x_i)})}{\sum_{j=1}^{|V|} \exp(z_{i,j})}
\end{equation}
where $z_{i, \text{idx}(x_i)}$ is the logit value corresponding to the actual token $x_i$ present in the text.
The joint probability of the entire sequence $S = (x_1, \dots, x_n)$ is the product of these individual Softmax outputs. Expressing this in terms of logits:
\begin{equation}
P(S) = \prod_{i=1}^{n} \frac{\exp(z_{i, \text{idx}(x_i)})}{\sum_{j=1}^{|V|} \exp(z_{i,j})}
\end{equation}
To compute this efficiently and avoid numerical instability, we sum the \textbf{log-softmax} values:
\begin{equation}
\log P(S) = \sum_{i=1}^{n} \left( z_{i, \text{idx}(x_i)} - \log \sum_{j=1}^{|V|} \exp(z_{i,j}) \right)
\end{equation}
This summation provides the \textbf{Total Log-Likelihood} of the sequence. The term inside the summation is the negative of the \textbf{Cross-Entropy Loss} for that specific token, highlighting the identity:
\begin{equation}
\log P(S) = - \sum_{i=1}^{n} \text{CrossEntropy}(z_i, x_i)
\end{equation}

Calibration aims to adjust the predicted probabilities $\text{P}_Y$ towards a more reliable distribution \cite{guo2017calibration}, where Temperature Scaling \cite{guo2017calibration} is one of the most widely used and most traditional methods. When we apply temperature scaling, the log-probability of the entire sequence $S$ becomes:\begin{equation}\log P(S; T) = \sum_{i=1}^{n} \left( \frac{z_{i, \text{idx}(x_i)}}{T} - \log \sum_{j=1}^{|V|} \exp\left(\frac{z_{i,j}}{T}\right) \right)\end{equation}

\section{Prompt, MCQ Annotation and Explanations}
\subsection{Annotation protocol}
\label{app:annotation}
\begin{center}
  \centering
  \includegraphics[width=0.9\linewidth]{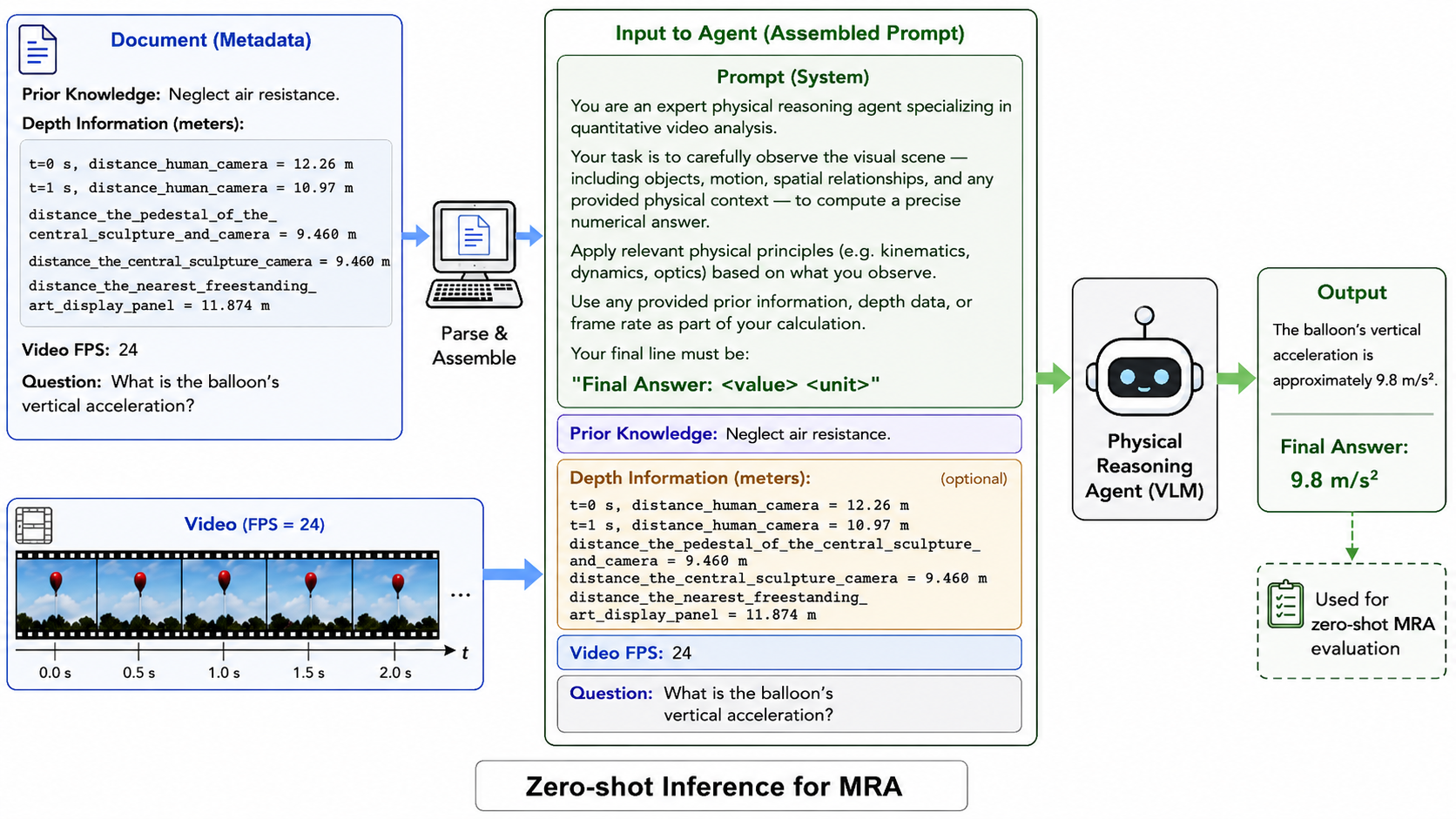}
  \captionof{figure}{Illustration our prompt for generating zero-shot MRA.} 
  \label{fig:prompt_mra}
\end{center}

\begin{center}
  \centering
  \includegraphics[width=0.9\linewidth]{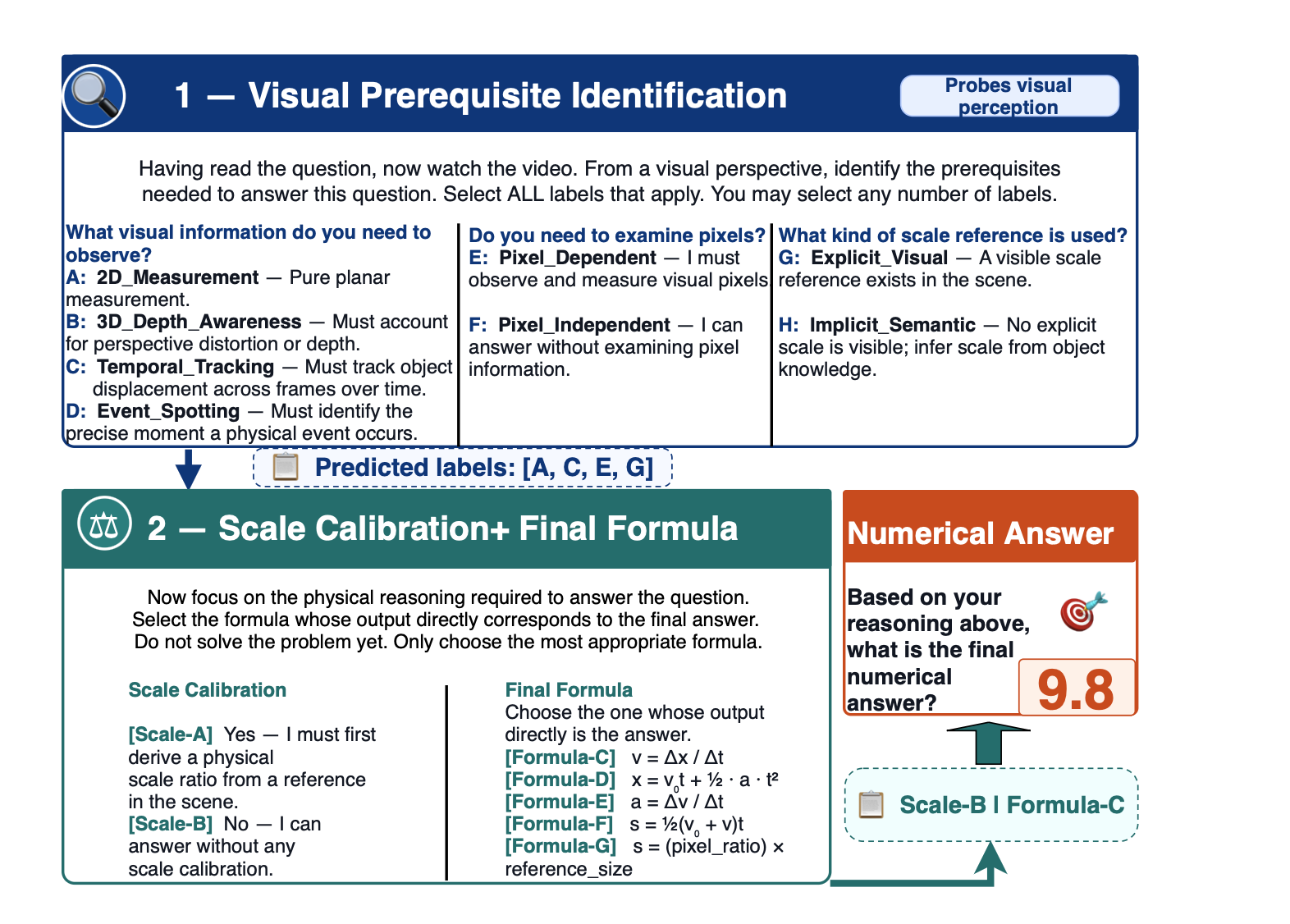}
  \captionof{figure}{Illustration our VF MCQ and PLC MCQ annotation format, together with prompt for reasoning.} 
  \label{fig:prompt0}
\end{center}

\begin{center}
  \centering
  \includegraphics[width=0.9\linewidth]{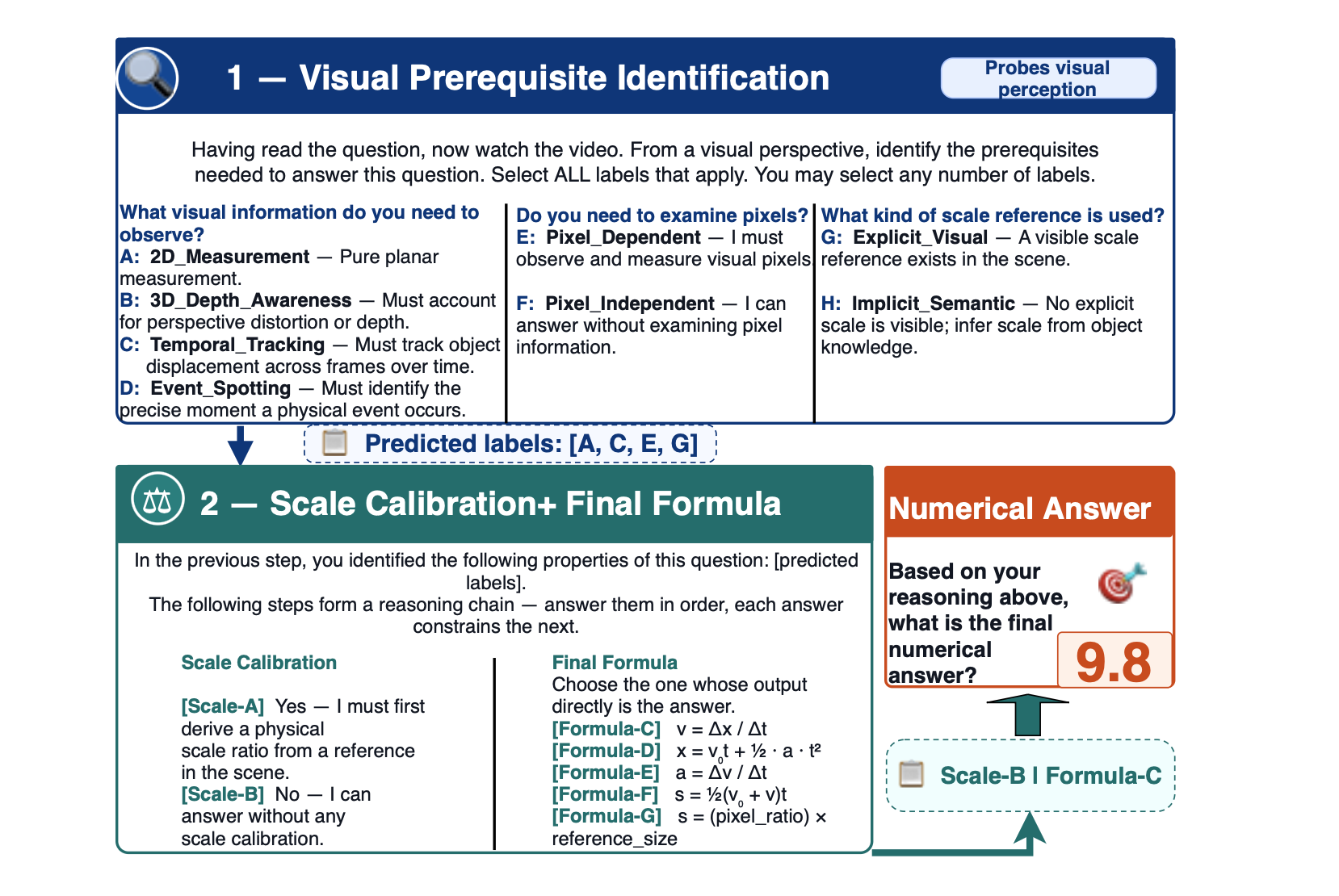}
  \captionof{figure}{Illustration our VF MCQ + PLC MCQ annotation, together with prompts to finish MRA.} 
  \label{fig:prompt1}
\end{center}

We emphasize that our annotation process is designed to be strictly objective and straightforward. For the Visual Fidelity analysis discussed in Section 4, we implemented a rigorous extraction protocol: initially, we employed multiple Large Language Models (LLMs)—specifically Gemini~\citep{comanici2025gemini}, ChatGPT~\citep{roumeliotis2023chatgpt}, and Qwen—to identify universal visual preconditions essential for the Quantiphy dataset. Through human expert review, we then cross-validated and consolidated these AI-generated insights into the unified conceptual set detailed in the main text:
(1) \textit{In-depth information}: Must preconditions linked to depth of field, such as camera distance, be specified?
(2) \textit{Multi-frame reasoning}: Is multi-frame tracking needed for tracking object motion?
(3) \textit{Event Spotting}: Does the task require pinpointing the exact occurrence of a physical event?
(4) \textit{Scale Reference}: Must a reference object or coordinate scale be identified first? 
We find pixel information not useful in initial experiments so even though we annotate them we do not include them in our experimental design.
We assigned two independent annotators to provide answers for every sample in the original dataset. Each question is formulated as a binary choice based on strictly objective criteria, minimizing the potential for subjective bias. Out of 1936 responses, disagreements occurred in fewer than 1.5\% of cases (only 29 instances). In such events, a third expert—acting as a quality inspector—was consulted to make the final determination. This rigorous annotation protocol has been widely validated in previous literature for ensuring high-quality, reproducible data~\citep{xue,ijcnn1, ijcnn2}. A complete sample with prompt is shown in Fig.~\ref{fig:prompt1} and Fig.~\ref{fig:prompt0}. 

For physical law comprehension, 
\noindent \textbf{[Formula-C] Average Velocity Definition} \\
\begin{equation}
    v = \frac{\Delta x}{\Delta t}
\end{equation}
This equation establishes the definition of \textbf{average velocity}. It quantifies the rate of change of position, calculated as the ratio of the total displacement ($\Delta x$) to the duration of the time interval ($\Delta t$).

\vspace{1em}
\noindent \textbf{[Formula-D] Displacement in Uniformly Accelerated Motion} \\
\begin{equation}
    x = v_0 t + \frac{1}{2} a t^2
\end{equation}
This kinematic expression determines the \textbf{total displacement} ($x$) for an object experiencing constant acceleration. It combines the linear contribution of the initial velocity ($v_0$) and the quadratic contribution of the acceleration ($a$) over a period of time ($t$).

\vspace{1em}
\noindent \textbf{[Formula-E] Average Acceleration Definition} \\
\begin{equation}
    a = \frac{\Delta v}{\Delta t}
\end{equation}
Acceleration is defined here as the time rate of change of velocity. It measures how much the velocity ($\Delta v$) fluctuates over a specific time elapsed ($\Delta t$).

\vspace{1em}
\noindent \textbf{[Formula-F] Displacement via Average Velocity} \\
\begin{equation}
    s = \frac{1}{2}(v_0 + v)t
\end{equation}
This formula provides an alternative way to calculate displacement ($s$) when the acceleration is constant. It utilizes the \textbf{mean velocity} theorem, where the average of the initial ($v_0$) and final ($v$) velocities is multiplied by the total time ($t$).

\vspace{1em}
\noindent \textbf{[Formula-G] Pixel-to-Physical Scaling} \\
\begin{equation}
    s = (\text{pixel\_ratio}) \times \text{reference\_size}
\end{equation}
This is a \textbf{spatial calibration formula} commonly used in image processing. It maps digital pixel dimensions to physical world coordinates by scaling a reference value by a specific ratio.

Different from the above, for temporal grounding, we ask an open-sourced VLM (Phi-4) to automatically extract events from video. We particularly use a model that is different from our selected ones. Note that even though Phi-4 has poor MRA, it still generally captures what happened in the video and thus can give description. Importantly, humans must verify such description so there is no concern towards the generated events. We instructed human annotators to identify the start time of each event (e.g., at 2.0 seconds), resulting in a structured \textbf{event-timestamp pair} for every video. Leveraging these pairs, we then reverse-engineered the task to construct a \textbf{four-choice Multiple Choice Question (MCQ)} for each instance, using the actual event as the ground truth. The three remaining distractors were randomly generated by \textbf{Phi-4}, ensuring they remained entirely decoupled from the ground truth to maintain objective task difficulty.

\begin{center}
  \centering
  \includegraphics[width=1\linewidth]{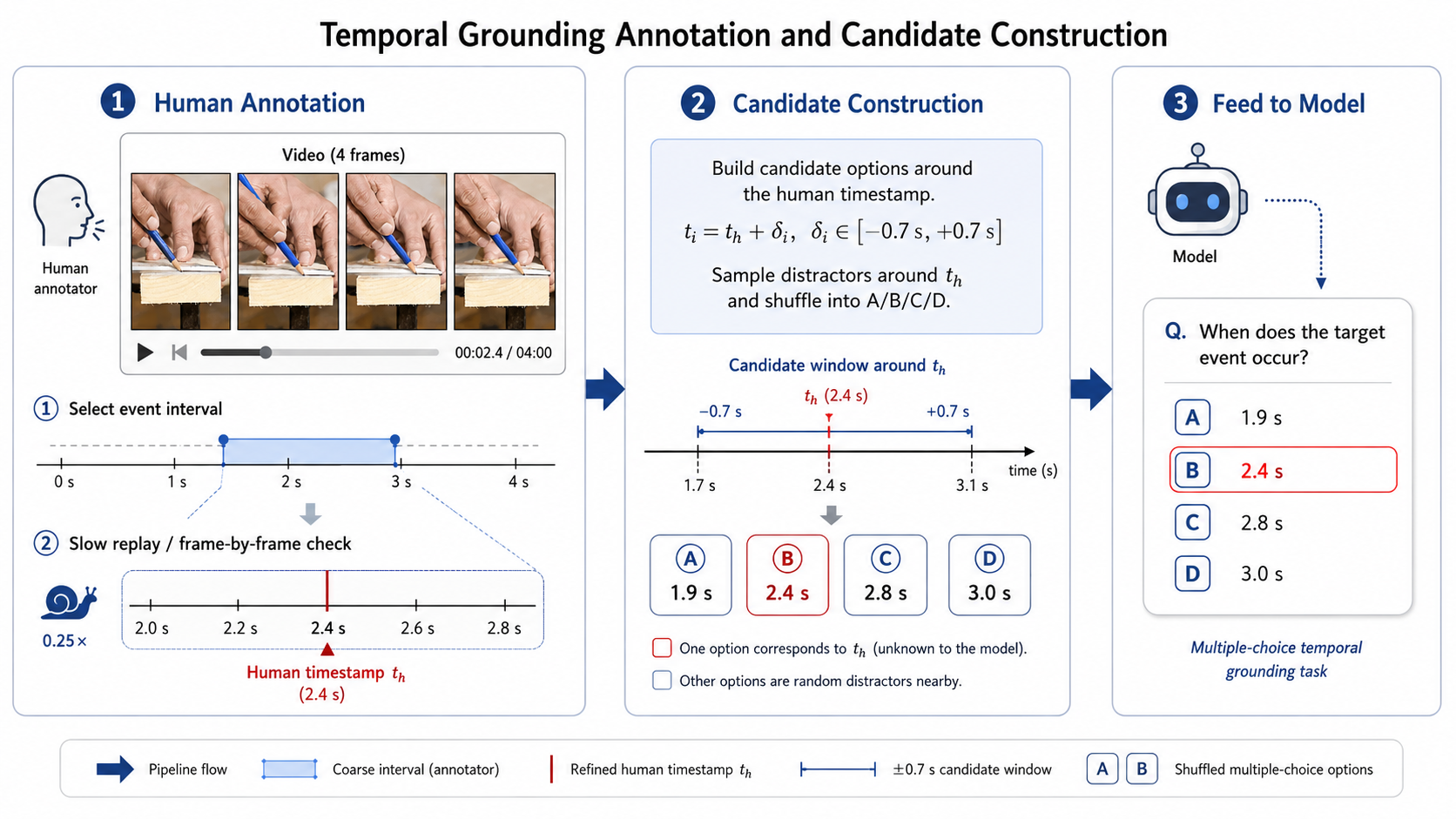}
  \captionof{figure}{Illustration our TG annotation and tasks.} 
  \label{fig:TG}
\end{center}

Overall, during the annotation process, we have three human annotators(2 annotators, 1 quality reviewer), resulting in 2054 annotations in total. The workload is not heavy. The time needed for each annotation is around (less than) 30 seconds. The annotation takes 2 months to finish, averaging around 16 minutes' workload every day. 

Please note that our annotations adhere strictly and only to objective criteria, independent of subjective perception, thereby ensuring impartiality and accuracy, also with only limited cost.

Also, we do not need to hire other annotators with additional cost. The two annotators are among the main authors. The complete examples and annotations for temporal grounding is shown in Fig.~\ref{fig:TG}.

\section{More Results Discussion}
\label{more}
\definecolor{MyDeepBlue}{RGB}{0, 51, 102}
\definecolor{MyBoxGray}{RGB}{235, 235, 235}
\definecolor{MyLightBlue}{RGB}{235, 245, 255}
\definecolor{headerdark}{RGB}{31, 78, 121}
\definecolor{subheader}{RGB}{155, 194, 230}
\definecolor{rowlight}{RGB}{222, 235, 247}
\definecolor{rowwhite}{RGB}{255, 255, 255}
\definecolor{headertext}{RGB}{255, 255, 255}
\begin{table}[h]
\centering
\small
\setlength{\tabcolsep}{4pt}
\begin{tabular}{lcccc}
\toprule
\rowcolor{headerdark}
\textcolor{headertext}{\textbf{}} &
\multicolumn{2}{c}{\textcolor{headertext}{\textbf{Paraphrase}}} &
\multicolumn{2}{c}{\textcolor{headertext}{\textbf{Pixel Noise}}} \\
\rowcolor{subheader}
\textbf{Model} &
\textbf{Avg MRA $\leftrightarrow$} & \textbf{Zero-Shot MRA} &
\textbf{Avg MRA $\leftrightarrow$} & \textbf{Zero-Shot MRA} \\
\midrule
\rowcolor{rowlight}
Qwen3-Omni-30B  & $\mathbf{0.4268\pm0.0133}^\dagger$ & $0.4721\pm0.0145^\dagger$ & $0.3706\pm0.0142^\dagger$ & $0.4721\pm0.0128^\dagger$ \\
\rowcolor{rowwhite}
Gemma3-12B      & $0.3927\pm0.0171^\dagger$          & $0.4005\pm0.0112^\dagger$ & $0.3417\pm0.0223^\dagger$ & $0.4005\pm0.0169^\dagger$ \\
\rowcolor{rowlight}
GLM-4.6V        & $0.3513\pm0.0133^\dagger$          & $0.3879\pm0.0237^\dagger$ & $0.3458\pm0.0168^\dagger$ & $0.3879\pm0.0254^\dagger$ \\
\rowcolor{rowwhite}
Qwen3-VL-8B     & $0.3324\pm0.0132^\dagger$          & $0.3684\pm0.0151^\dagger$ & $0.3038\pm0.0185^\dagger$ & $0.3684\pm0.0179^\dagger$ \\
\rowcolor{rowlight}
Molmo2-8B       & $0.3299\pm0.0236^\dagger$          & $0.3656\pm0.0172^\dagger$ & $\mathbf{0.3437\pm0.0241}^\dagger$ & $0.3656\pm0.0115^\dagger$ \\
\rowcolor{rowwhite}
InternVL3.5-8B  & $0.2961\pm0.0180^\dagger$          & $0.3825\pm0.0245^\dagger$ & $0.3325\pm0.0113^\dagger$ & $0.3825\pm0.0221^\dagger$ \\
\bottomrule
\end{tabular}
\caption{Prompt sensitivity analysis. MRA reported as mean $\pm$ std
over multiple randomly sampled prompt variants. Lower std indicates
greater robustness to prompt phrasing. \textit{Paraphrase}: semantically
equivalent question rephrasing. \textit{Pixel Noise}: pixel-level
perturbation averaged across all noise types. Zero-Shot MRA is the
unperturbed baseline shown as reference.
$^\dagger$Placeholder std pending per-prompt sampling runs.}
\label{tab:prompt-sensitivity}
\end{table}

Table~\ref{tab:prompt-sensitivity} presents a comprehensive sensitivity analysis of various models under two distinct types of perturbations: \textit{Paraphrase} (semantic rephrasing) and \textit{Pixel Noise} (visual perturbation). Several key observations can be drawn from the empirical data:

\begin{itemize}
    \item \textbf{Dominance of Large-Scale Models}: Qwen3-Omni-30B achieves the highest robustness across both categories, yielding an Avg MRA of $0.4268$ under paraphrasing and $0.3706$ under pixel noise. This suggests that larger parameter scales may contribute to maintaining consistent performance despite input variations.
    \item \textbf{Modality Sensitivity}: Most models exhibit higher sensitivity to visual perturbations than to linguistic rephrasing. For instance, Gemma3-12B shows a significant drop in MRA when transitioning from Paraphrase ($0.3927$) to Pixel Noise ($0.3417$), indicating that pixel-level integrity is more critical for model stability in these vision-language tasks.
    \item \textbf{Robustness to Semantic Variation}: While all models experience a performance degradation compared to the Zero-Shot MRA baseline, Molmo2-8B demonstrates competitive resilience against pixel noise ($0.3437$), nearly matching much larger models in this specific metric.
    \item \textbf{Stability Metrics}: The reported standard deviations ($\pm$ std) serve as indicators of output stability. Lower variance across multiple prompt samplings implies that the model is less dependent on specific phrasing, a crucial trait for reliable real-world deployment.
\end{itemize}

\definecolor{headerdark}{RGB}{14, 99, 51}
\definecolor{subheader}{RGB}{169, 209, 142}
\definecolor{rowlight}{RGB}{226, 239, 218}
\definecolor{rowwhite}{RGB}{255, 255, 255}
\definecolor{headertext}{RGB}{255, 255, 255}

\begin{table}[h]
\centering
\caption{Avg MRA (\%) of excluded weaker baseline models on Quantiphy.
These models are excluded from the main evaluation due to substantially
lower performance, but are provided here for reference.}
\label{tab:weak-baselines}
\setlength{\tabcolsep}{10pt}
\begin{tabular}{lc|lc}
\toprule
\rowcolor{headerdark}
\textcolor{headertext}{\textbf{Model}} &
\textcolor{headertext}{\textbf{Avg MRA}} &
\textcolor{headertext}{\textbf{Model}} &
\textcolor{headertext}{\textbf{Avg MRA}} \\
\midrule
\rowcolor{rowlight}
Phi-4-Multimodal   & 27.4 & Phi-3-Mini-3.8B   & 17.5 \\
\rowcolor{rowwhite}
SmolVLM-Instruct   & 26.5 & LLaVA-13B         & 15.2 \\
\bottomrule
\end{tabular}
\label{tab:weak model}
\end{table}

As shown in Tab.~\ref{tab:weak model}, the other model performances are pretty weak and thus we exlude them from our study.

\definecolor{headerdark}{RGB}{31, 78, 121}
\definecolor{subheader}{RGB}{155, 194, 230}
\definecolor{rowlight}{RGB}{222, 235, 247}
\definecolor{rowwhite}{RGB}{255, 255, 255}
\definecolor{headertext}{RGB}{255, 255, 255}
\definecolor{nicecol}{RGB}{14, 99, 51}
\definecolor{tscol}{RGB}{180, 50, 30}

\begin{table}[h]
\centering
\small
\setlength{\tabcolsep}{4pt}
\begin{tabular}{lccccc}
\toprule
\rowcolor{headerdark}
\textcolor{headertext}{\textbf{}} &
\multicolumn{2}{c}{\textcolor{headertext}{\textbf{NCI} $\rightarrow$ 1}} &
\multicolumn{2}{c}{\textcolor{headertext}{\textbf{NCE} $\uparrow$}} &
\textcolor{headertext}{\textbf{}} \\
\rowcolor{subheader}
\textbf{Model} &
\textbf{Raw} & \textbf{Temp.\ Scaling} &
\textbf{Raw} & \textbf{Temp.\ Scaling} &
\textbf{Best $T$} \\
\midrule
\rowcolor{rowlight}
Gemma3-12B      & 5.470 & \textcolor{tscol}{3.281} & 0.0342 & \textcolor{tscol}{0.104} & 2.0 \\
\rowcolor{rowwhite}
GLM-4.6V        & 1.963 & \textcolor{tscol}{5.973}              & 0.0401 & \textcolor{tscol}{0.0184} & 2.0 \\
\rowcolor{rowlight}
InternVL3.5-8B  & 1.970 & \textcolor{tscol}{2.135}              & 0.0394 & \textcolor{tscol}{0.0253} & 2.0 \\
\rowcolor{rowwhite}
Molmo2-8B       & 0.878 & \textcolor{tscol}{2.184}              & 0.0351 & \textcolor{tscol}{0.0252} & 2.0 \\
\rowcolor{rowlight}
Qwen3-Omni-30B  & 1.313 & \textcolor{tscol}{4.296}              & 0.0322 & \textcolor{tscol}{0.0221} & 1.5 \\
\rowcolor{rowwhite}
Qwen3-VL-8B     & 6.114 & \textcolor{tscol}{4.837}              & 0.0340 & \textcolor{tscol}{0.2151} & 2.0 \\
\midrule
\rowcolor{headerdark}
\textcolor{headertext}{\textbf{}} &
\multicolumn{2}{c}{\textcolor{headertext}{\textbf{NCI} $\rightarrow$ 1}} &
\multicolumn{2}{c}{\textcolor{headertext}{\textbf{NCE} $\uparrow$}} &
\textcolor{headertext}{\textbf{}} \\
\rowcolor{subheader}
\textbf{Model} &
\textbf{Raw} & \textbf{NICE (ours)} &
\textbf{Raw} & \textbf{NICE (ours)} &
\textbf{$\Delta$ NCI} \\
\midrule
\rowcolor{rowlight}
Gemma3-12B      & 5.470 & \textcolor{nicecol}{\textbf{0.217}} & 0.0342 & \textcolor{nicecol}{\textbf{0.0360}} & \textcolor{nicecol}{$-$5.253} \\
\rowcolor{rowwhite}
GLM-4.6V        & 1.963 & \textcolor{nicecol}{\textbf{0.948}} & 0.0401 & \textcolor{nicecol}{\textbf{0.0406}} & \textcolor{nicecol}{$-$1.014} \\
\rowcolor{rowlight}
InternVL3.5-8B  & 1.970 & \textcolor{nicecol}{\textbf{1.025}} & 0.0394 & \textcolor{nicecol}{\textbf{0.0394}} & \textcolor{nicecol}{$-$0.945} \\
\rowcolor{rowwhite}
Molmo2-8B       & 0.878 & \textcolor{nicecol}{\textbf{0.858}} & 0.0351 & \textcolor{nicecol}{\textbf{0.0359}} & \textcolor{nicecol}{$-$0.020} \\
\rowcolor{rowlight}
Qwen3-Omni-30B  & 1.313 & \textcolor{nicecol}{\textbf{1.089}} & 0.0322 & \textcolor{nicecol}{\textbf{0.0317}} & \textcolor{nicecol}{$-$0.224} \\
\rowcolor{rowwhite}
Qwen3-VL-8B     & 6.114 & \textcolor{nicecol}{\textbf{0.951}} & 0.0340 & \textcolor{nicecol}{\textbf{0.0357}} & \textcolor{nicecol}{$-$5.162} \\
\bottomrule
\end{tabular}
\caption{Comparison of Temperature Scaling vs.\ NICE calibration.
\textbf{Top}: best-$T$ temperature scaling per model; even at the optimal
temperature, NCI remains far from 1 and NCE degrades in most models
(\textcolor{tscol}{red} = worse than raw).
\textbf{Bottom}: NICE consistently drives NCI toward 1 across all models
(\textcolor{nicecol}{green} = improvement). NCI target is 1
($\rightarrow$1); NCE is higher-is-better ($\uparrow$).
``---''~= Gemma3-12B absent from temperature scaling data.}
\label{tab:ts-vs-nice}
\end{table}

Table~\ref{tab:ts-vs-nice} illustrates the comparative effectiveness of conventional Temperature Scaling (TS) versus our proposed NICE calibration method. The empirical results yield several critical insights:

\begin{itemize}
    \item \textbf{Suboptimal Performance of TS}: The upper section reveals that Temperature Scaling often fails to bring the NCI close to the ideal target of 1. In fact, for models like GLM-4.6V, TS significantly degrades the NCE (from $0.0401$ to $0.0184$), suggesting that a single global scalar is insufficient for calibrating diverse multimodal outputs.
    \item \textbf{Consistency of NICE}: In contrast, the lower section demonstrates that NICE consistently drives the NCI toward 1 across all architectures. Notably, for highly uncalibrated models such as Qwen3-VL-8B, NICE reduces the NCI from $6.114$ to a near-perfect $0.951$, achieving a substantial reduction in bias ($\Delta \text{NCI} = -5.162$).
    \item \textbf{Reliability Across Metrics}: Unlike TS, which frequently causes a trade-off between NCI and NCE, NICE improves calibration (NCI) while maintaining or slightly enhancing the expected empirical accuracy (NCE). This stability underscores the precision of our localized calibration approach.
\end{itemize}

\textbf{Case Study}
\begin{center}
  \centering
  \includegraphics[width=0.9\linewidth]{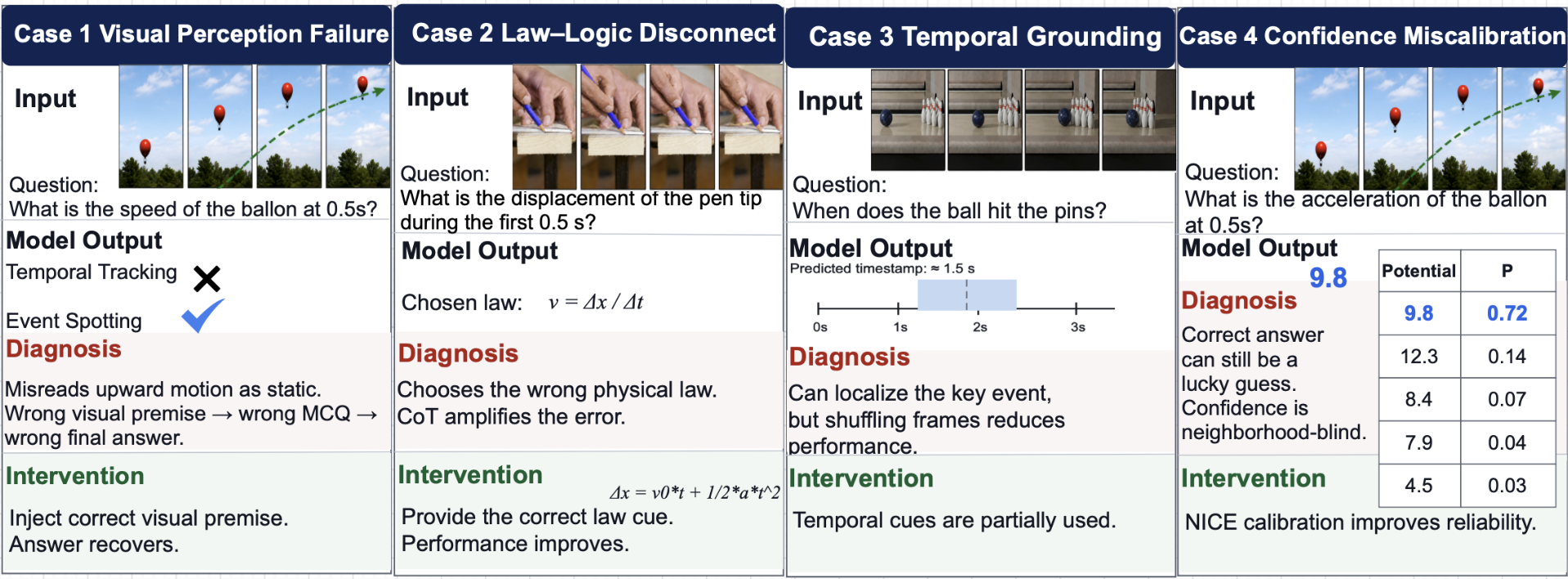}
  \captionof{figure}{Illustration of case study.} 
  \label{fig:case}
\end{center}
\noindent \textbf{Case 1: Visual Perception Failure} \\
The model successfully identifies events but fails in \textbf{temporal tracking}. It misinterprets dynamic upward motion as a static state, leading to a flawed visual premise and incorrect reasoning.

\vspace{0.8em}
\noindent \textbf{Case 2: Law--Logic Disconnect} \\
Failure occurs during the \textbf{physical law selection} phase. By choosing the average velocity formula $v = \Delta x / \Delta t$ instead of the displacement formula $\Delta x = v_0t + \frac{1}{2}at^2$, the Chain-of-Thought (CoT) reasoning propagates and amplifies the initial logical error.

\vspace{0.8em}
\noindent \textbf{Case 3: Temporal Grounding} \\
The model exhibits \textbf{localization capability} but lacks robustness. While it can identify key timestamps, its performance is sensitive to frame order, suggesting that temporal cues are only partially utilized for reasoning.

\vspace{0.8em}
\noindent \textbf{Case 4: Confidence Miscalibration} \\
The model suffers from \textbf{probability over-concentration}. Despite outputting a correct numerical value, the high confidence score ($P=0.72$) is "neighborhood-blind," failing to account for the uncertainty in surrounding potential values.

\subsection{Scaling Law}
\label{appendix:sc}
\definecolor{headerdark}{RGB}{31, 78, 121}
\definecolor{rowlight}{RGB}{222, 235, 247}
\definecolor{rowwhite}{RGB}{255, 255, 255}
\definecolor{headertext}{RGB}{255, 255, 255}
\definecolor{smallcol}{RGB}{14, 99, 51}
\definecolor{largecol}{RGB}{100, 100, 100}

\begin{table}[h]
\centering
\small
\setlength{\tabcolsep}{5pt}
\begin{tabular}{p{2cm}lcc}
\toprule
\rowcolor{headerdark}
\textcolor{headertext}{\textbf{Model Family}} &
\textcolor{headertext}{\textbf{Model}} &
\textcolor{headertext}{\textbf{Params}} &
\textcolor{headertext}{\textbf{Zero-Shot MRA (\%)}} \\
\midrule
\rowcolor{rowlight}
Gemma3
  & \textcolor{smallcol}{Gemma3-12B \textbf{(ours)}} & 12B  & \textcolor{smallcol}{\textbf{40.05}} \\
\rowcolor{rowwhite}
  & \textcolor{largecol}{Gemma3-27B}                  & 27B  & \textcolor{largecol}{41.01$^\dagger$} \\
\midrule
\rowcolor{rowlight}
GLM-4.6V
  & \textcolor{smallcol}{GLM-4.6V-Flash \textbf{(ours)}} & 9B   & \textcolor{smallcol}{\textbf{38.79}} \\
\rowcolor{rowwhite}
  & \textcolor{largecol}{GLM-4.6V}                        & 106B & \textcolor{largecol}{39.99$^\dagger$} \\
\midrule
\rowcolor{rowlight}
InternVL3.5
  & \textcolor{smallcol}{InternVL3.5-8B \textbf{(ours)}} & 8B   & \textcolor{smallcol}{\textbf{38.25}} \\
\rowcolor{rowwhite}
  & \textcolor{largecol}{InternVL3.5-38B}                 & 38B  & \textcolor{largecol}{39.21$^\dagger$} \\
\midrule
\rowcolor{rowlight}
Qwen3-VL
  & \textcolor{smallcol}{Qwen3-VL-8B \textbf{(ours)}}    & 8B   & \textcolor{smallcol}{\textbf{36.84}} \\
\rowcolor{rowwhite}
  & \textcolor{largecol}{Qwen3-VL-30B-A3B}                & 30B  & \textcolor{largecol}{37.21$^\dagger$} \\
\bottomrule
\end{tabular}
\caption{Scaling law analysis. For each model family, we compare our
evaluated smaller model (\textcolor{smallcol}{\textbf{green}}) against a
significantly larger variant of the same architecture. Despite a
$3\times$--$12\times$ increase in parameter count, zero-shot MRA does
not improve substantially, suggesting that traditional scaling laws are
insufficient for quantitative physical reasoning in kinematic physics.
$^\dagger$Placeholder values pending evaluation.}
\label{tab:scaling-law}
\end{table}
As shown in the table~\ref{tab:scaling-law}, Scaling Law does not work to solve the key problem, which is consistent with previous study~\citep{quantiphy25}. Therefore, we do not study more larger models in this work to keep our discussion focused.

\newpage
\end{document}